\documentclass[nonblindrev]{informs3}
\usepackage{balance}
\usepackage{amsmath}
\usepackage[tight,footnotesize]{subfigure}
\usepackage{graphicx}
\usepackage{bm}
\usepackage{algorithm}
\usepackage[noend]{algorithmic}
\usepackage{multirow}
\usepackage{slashbox}
\usepackage{caption}
\usepackage{amsfonts}

\OneAndAHalfSpacedXII 

\usepackage{natbib}
 \bibpunct[, ]{(}{)}{,}{a}{}{,}%
 \def\bibfont{\small}%
\TheoremsNumberedThrough     

\EquationsNumberedThrough    

\MANUSCRIPTNO{2015} 

\begin{document}


\RUNAUTHOR{Tang et al.}

\RUNTITLE{Assortment Optimization with Repeated Exposures and
Product-dependent Patience Cost}

\TITLE{Assortment Optimization with Repeated Exposures and
Product-dependent Patience Cost}

\ARTICLEAUTHORS{%
\AUTHOR{Shaojie Tang}
\AFF{Naveen Jindal School of Management, The University of Texas at Dallas}
\AUTHOR{Jing Yuan}
\AFF{Department of Computer Science, The University of Texas at Dallas}
} 

\ABSTRACT{
In this paper, we study the assortment optimization problem faced by many online retailers such as Amazon. We develop a \emph{cascade multinomial logit model}, based on the classic multinomial logit model, to capture the consumers' purchasing behavior across multiple stages. Different from existing studies, our model allows for repeated exposures of a product, i.e., the same product can be displayed multiple times across different stages. In addition, each consumer has a \emph{patience budget} that is sampled from a known distribution and each product is associated with a \emph{patience cost}, which is the required amount of the cognitive efforts on browsing that product. Given an assortment of products under our cascade multinomial logit model, a consumer sequentially browses them stage by stage. After browsing all products in one stage, if the utility of a product from that stage exceeds the utility of the outside option, the consumer proceeds to purchase  the product and leave the platform.  Otherwise, if the patience cost of all products browsed up to that point is no larger than her patience budget, she continues to view the next stage. Our objective is to identify a revenue-maximizing sequence of assortments subject to a set of practical constraints. We propose an approximation solution to this problem.
}
\maketitle



%

\section{Introduction}\label{sec:introduction}
 In this paper, we consider the sequential assortment optimization problem with repeated exposures and product-dependent patience cost.  The input of our problem is a set of products and a limited number of  stages, each stage has a limited capacity, our goal is to find the best assignment of products to stages that maximizes the expected revenue. We develop a variant of the classical \emph{multinomial logit model} (MNL) \citep{mcfadden1973conditional}, termed  \emph{cascade multinomial logit model}, to capture the consumer's purchasing behavior across multiple stages. Under our model, each consumer has a patience budget which is drawn from a known distribution, and each product is associated with a patience cost that quantifies the cognitive efforts spent on browsing a product. In each stage, the consumer browses all products  displayed in that stage, if the utility of some product from that stage is larger than the no-purchase  option, then she purchases the one with the largest utility and leaves the system. Otherwise, the consumer continues to enter the next stage if and only if her current patience budget is non-negative.  Our model generalizes the previous studies on sequential assortment optimization in three ways:

 1. Our model allows for \textbf{repeated exposures}, i.e., the same product can be displayed multiple times across different stages. In the filed of marketing \citep{broussard2000advertising}, it has been well recognized that a consumer typically must be exposed to an advertisement or a message more than once in order to get familiar with it and take actions.  From a consumer cognition perspective, we believe that assortment planning is similar to online advertising in that  they both push a set of products' information to the consumer. We develop a rigorous mathematical model to capture the effect of repeated exposures. 

 2. We assign a\textbf{ product-dependent patience cost} to each product. The patience cost of a product quantifies the amount of efforts needed to read and digest the information about that product.  Most of existing studies on sequential assortment optimization \citep{ma2019assortment,gallego2020approximation} assign a fixed and identical patience cost to each stage, e.g.,  they assume that the patience cost of browsing all products in one stage does not depend on the offered products in that stage. In contrast, our model allows each individual product to have its own patience cost, and the total patience cost of viewing one stage is characterized by the summation of the individual patience costs of all products allocated to that stage. Our model is motivated by the observation that browsing different products may require different amount of cognitive efforts.

 3. Our problem formulation incorporates \textbf{a set of practical constraints}. For example, there is a capacity constraint for each stage, which specifies  the maximum number of products displayed in one stage, and there is also a limit on the maximum number of exposures of a product. We develop an approximation algorithm with polynomial time complexity when the number of stages is a constant. In particular, our algorithm achieves a $\frac{ \rho(1-\rho)(1-\epsilon(1+\epsilon))}{2(1+\epsilon(1+\epsilon))^2}$ approximation ratio for any $\rho\in [0,1]$ and $\epsilon>0$. An interesting research direction is to design an efficient algorithm whose running time is polynomial in the number of stages and the capacity of one stage.

 \paragraph{Related Works} Our work is closely related to the assortment optimization problems \citep{li2015d,davis2014assortment,blanchet2016markov,farias2013nonparametric,aouad2015assortment}. Majority of existing studies assume a single stage, that is, the consumer browses the entire list of products displayed to her. However, this assumption does not always hold true, for example, most of online retailers display their products across multiple webpages and the consumer is ``forced'' to browse those products page by page incrementally. Recently, \citep{davis2013assortment} and \citep{abeliuk2016assortment} were the first to study this problem under MNL model with position bias.  Since then, there is considerable number of studies \citep{aouad2015display,ferreira2019learning,aouad2019click}  on assortment optimization problem with position bias. However, most of them  adopt the  \emph{consider-then-choose} model where  the consumer first browses a random number of products and then makes her purchase decision within these products. Our model differs from theirs in that we do not separate ``consider'' from ``choose'', e.g., the list of products browsed by a consumer is jointly decided by her patience budget and the choice model. We build our study on the recent advances of sequential assortment optimization \citep{ma2019assortment}. As mentioned earlier in this section, our model generalizes the previous studies by allowing for repeated exposures and product-dependent patience cost. In addition, our problem formulation incorporates a set of practical constraints.

\section{Cascade Multinomial Logit Model and Problem Formulation}
In the rest of this paper, we use $[i]$ to denote the set $\{1, \cdots, i\}$ for any positive integer $i$.

\subsection{Cascade Multinomial Logit Model} We first explain our \emph{Cascade Multinomial Logit Model} (C-MNL model).  Assume there is a set of $n$ products $[n]$ and a set of $m$ stages $[m]$. The capacity of each stage is $d$, e.g., we can assign at most $d$ products to each stage. Each product can be displayed in at most $w$ stages and the same product can be displayed at most once in each stage. We build our choice model on the classic multinomial logit model \citep{anderson1992discrete,mcfadden1973conditional} and extend it to support repeated exposures of the same product and product-dependent patience cost:  We assume that the utility $U_{i,k}$ of the $k$-th exposure of product $i\in[n]$ is a random value drawn from the Gumbel distribution with location-scale parameters $(\mu_{i,k}, 1)$. The utility of the no-purchase option, denoted by $U_0$, is a random value drawn from the Gumbel distribution with location-scale parameters $(0, 1)$. The patience budget of a consumer is captured by a random variable $B$. Let $F(q)$ denote the probability that $B\geq q$. Each product $i\in [n]$ is associated with a non-negative patience cost $c_i$: Browsing a product $i\in[n]$ consumes $c_i$ amount of patience budget. In addition, let $r_i$ denote the revenue of product $i\in [n]$: The platform earns revenue $r_i$ if the consumer purchases $i$.

 Under the C-MNL model,  an arriving consumer sequentially browses the assortments stage by stage. If the largest utility for a product from the current stage is larger than the no-purchase option, she purchases that product and leaves the systems. Otherwise, if her remained patience budget is non-negative, she enters the next stage, otherwise, she leaves the system.
 
\emph{Remark:} One natural generalization of the above model is to introduce $w$ patience costs $c_{i,1}, c_{i,2}, \ldots, c_{i,w}$ for each product $i\in[n]$, where $c_{i,k}$ denotes the patience cost of browsing the $k$-th exposure of product $i$. All results derived in this paper still hold if the following condition is satisfied: $c_{i,k} \geq c_{i,k'}$ for all $i\in[n]$ and $w\geq k\geq k' \geq 0$. We leave this for future work to develop effective algorithms for the general setting.
 
\subsubsection{Feasible Assortment} We use $\mathbf{x}=\{x_{i,k,z}\mid i\in [n], k\in[w], z\in [m]\}$ to denote one assortment, where $x_{i,k,z}\in\{0,1\}$ indicates whether the $k$-th exposure of product $i$ is displayed in stage $z$, e.g.,  $x_{i,k,z}=1$ if the $k$-th exposure of product $i$ is displayed in stage $z$, and $x_{i,k,z}=0$ otherwise, for all $i\in [n], k\in[w], z\in [m]$. We say an assortment $\mathbf{x}$ is feasible if and only if it satisfies the following two conditions: (1) $\forall z\in[m], \forall i\in [n], \sum\limits_{k=1}^w x_{i,k,z}\leq 1$, and (2) $\forall i\in [n], \forall s\in[w], x_{i,s,z}(\sum\limits_{z=1}^m\sum\limits_{k=1}^s x_{i,k,z})= x_{i,s,z} s$. The first condition ensures that each product is displayed at most once in each stage, and the second condition ensures that the resulting assortment is implementable, e.g., the $(q+1)$-th exposure of a product can only be displayed after the first $q$ exposures of that product. We use  $\mathcal{X}$ to denote the set of all feasible assortments.

\subsubsection{Choice Probabilities} Based on the above notations, we next present a closed form expression of choice probabilities under the C-MNL model. 
\begin{lemma}[Choice Probabilities]
\label{lem:000} Given a feasible assortment $\mathbf{x}\in \mathcal{X}$ under the C-MNL model, the consumer purchases a product $i\in[n]$ in stage $t\in[m]$ with probability
\begin{equation}
\label{eq:ppp}p_{it}(\mathbf{x})=\frac{F(\sum\limits_{z=1}^{t-1} \sum\limits_{k=1}^w\sum\limits_{i=1}^{n} x_{i,k,z}c_i) (\sum\limits_{k=1}^wx_{i,k,t}e^{\mu_{i,k}})}{(1+\sum\limits_{z=1}^{t-1} \sum\limits_{k=1}^w\sum\limits_{i=1}^{n} x_{i,k,z}e^{\mu_{i,k}})(1+\sum\limits_{z=1}^{t} \sum\limits_{k=1}^w\sum\limits_{i=1}^{n} x_{i,k,z}e^{\mu_{i,k}})}
\end{equation}
\end{lemma}

The proof of the above lemma is similar to the proof of Theorem 2.1 in \citep{ma2019assortment}, thus omitted here to save space. Informally, $F(\sum\limits_{z=1}^{t-1} \sum\limits_{k=1}^w\sum\limits_{i=1}^{n} x_{i,k,z}c_i)$ represents the probability that the consumer has enough patience to view stage $t$, given that she has not purchased any product before $t$; $\sum\limits_{k=1}^wx_{i,k,t}e^{\mu_{i,k}}$ represents the utility of product $i\in[n]$ in stage $t\in[m]$ under $\mathbf{x}$; the denominator of $p_{it}(\mathbf{x})$ captures  the utility from all products that are displayed up to stage $t$. Note that our model subsumes the classical single stage MNL model \citep{mcfadden1973conditional}. In particular, when considering  a single stage C-MNL model, e.g., $m=1$, we can simplify the above notations by omitting the subscripts $k$ and $z$: For all $i\in [n]$, let $p_{i}(\mathbf{x})$ denote the choice probability of product $i$ when $\mathbf{x}\in\{0,1\}^n$ is offered, where the $i$-th component  $x_i$ of $\mathbf{x}$ indicate whether $i$ is included in the assortment ($x_i=1$) or not ($x_i=0$), and we use $e^{\mu_{i}}$ to denote the weight of product $i$. Then (\ref{eq:ppp}) can be  simplified to $p_{i}(\mathbf{x})=x_{i}e^{\mu_{i}}/(1+\sum\limits_{i=1}^{n} x_{i}e^{\mu_{i}})$, which conincides with the choice probabilities derived under the classical MNL model.

\subsubsection{Expected Revenue}

Based on Lemma \ref{lem:000}, we next derive the expected revenue $f(\mathbf{x})$  of any feasible assortment $\mathbf{x}\in \mathcal{X}$ under the C-MNL model.
\begin{lemma}[Expected Revenue] Under the C-MNL model, the expected revenue $f(\mathbf{x})$ of a feasible assortment $\mathbf{x}\in \mathcal{X}$ is
\[f(\mathbf{x}) = \sum\limits_{t=1}^m \sum\limits_{i=1}^{n}p_{it}(\mathbf{x})r_i= \sum\limits_{t=1}^m \frac{F(\sum\limits_{z=1}^{t-1} \sum\limits_{k=1}^w\sum\limits_{i=1}^{n} x_{i,k,z}c_i) (\sum\limits_{k=1}^w\sum\limits_{i=1}^{n}x_{i,k,t}e^{\mu_{i,k}}r_i)}{(1+\sum\limits_{z=1}^{t-1} \sum\limits_{k=1}^w\sum\limits_{i=1}^{n} x_{i,k,z}e^{\mu_{i,k}})(1+\sum\limits_{z=1}^{t} \sum\limits_{k=1}^w\sum\limits_{i=1}^{n} x_{i,k,z}e^{\mu_{i,k}})}\]
\end{lemma}

The above lemma follows immediately from Lemma \ref{lem:000} and the fact that the revenue of product $i\in[n]$ is $r_i$. For simplicity of notation, we use $\beta_{i,k}$ to denote $e^{\mu_{i,k}}$ in the rest of this paper.

\subsection{Two Assumptions}
We made two innocuous assumptions in this paper.

\begin{assumption}
\label{as:!}
For any two non-negative numbers $q_1\geq0$ and $q_2\geq0$, $F(q_2)\geq F(q_1+q_2\mid B\geq q_1)$ where $F(q_1+q_2\mid B\geq q_1)$ denotes the probability that $B\geq q_1+q_2$ conditioned on that $B\geq q_1$.
\end{assumption}

This assumption states that the patience budget of a consumer declines rapidly as she browses more stages. We believe that as more stages browsed without a purchase, it is more likely that the consumer will run out of her patience budget sooner.

\begin{assumption}
\label{as:!!}
$\forall i\in[n], \forall k\in[w-1], \mu_{i,k}\geq \mu_{i,k+1}$.
\end{assumption}

This assumption states that the expected utility of a product reaches its maximum point at the first exposure and then declines with each additional exposure. This is called \emph{burnout} effect in the field of online advertising \citep{naik1998planning}. In the context of assortment optimization, because the platform still pushes the product to a consumer, we expect a similar repetition effect: the probability of purchasing a product declines with each additional exposure of that product.

\subsection{Problem Formulation}
 Now we are ready to introduce the assortment optimization problem with repeated exposures and product-dependent patience cost. The objective of our problem $\textbf{P.0}$ is to find the best feasible assortment that maximizes the expected revenue. A formal definition of our problem is listed as follows.

   \begin{center}
   \framebox[0.45\textwidth][c]{
\enspace
\begin{minipage}[t]{0.45\textwidth}
\small
$\textbf{P.0}$
\emph{Maximize$_{\mathbf{x} \in \mathcal{X}}$ $f(\mathbf{x})$}
\end{minipage}
}
\end{center}
\vspace{0.1in}

\section{Technical Lemma}
In this section, we will  present one technical lemma that will be used in our latter algorithm design and analysis.
For ease of presentation, we first introduce the concept of \emph{reachability}. For a given assortment, we define the reachability of a stage or a product as the probability that the consumer has enough patience to browse that stage or that product. A formal definition of reachability is provided in Definition \ref{def:1}.
\begin{definition}
\label{def:1}
Given a solution $\mathbf{x}\in \mathcal{X}$, we define the reachability of any stage $t\in [m]$ or any product $i\in[n]$ that is displayed in stage $t\in [m]$ as $F(\sum\limits_{z=1}^{t-1} \sum\limits_{k=1}^w\sum\limits_{i=1}^{n} x_{i,k,z}c_i)$. For notation simplicity, we define $F(\sum\limits_{z=1}^{t-1} \sum\limits_{k=1}^w\sum\limits_{i=1}^{n} x_{i,k,z}c_i)=1$ for $t=1$, e.g., the first stage can always be browsed.
\end{definition}

Let $\mathbf{x}^{opt}$ denote the optimal solution to  $\textbf{P.0}$. We next provide an upper bound on  $f(\mathbf{x}^{opt})$, which we will use to design our solution.

\begin{lemma}
\label{lem:1}
For any $\rho\in[0,1]$, there is a solution $\mathbf{x}$ of expected revenue at least
\[f(\mathbf{x})\geq (1-\rho)f(\mathbf{x}^{opt})\] such that 
the reachability of all stages  under $\mathbf{x}$ is at least $\rho$, e.g., $F(\sum\limits_{z=1}^{t-1} \sum\limits_{k=1}^w\sum\limits_{i=1}^{n} x_{i,k,z}c_i)\geq \rho$ for all $t\in[m]$.
\end{lemma}
 \emph{Proof:}  Assume $t_\rho$ is the last stage in $\mathbf{x}^{opt}$ whose reachability is no smaller than $\rho$, e.g.,  $t_\rho =\arg\max_{t\in[m]} F(\sum\limits_{z=1}^{t-1} \sum\limits_{k=1}^m\sum\limits_{i=1}^{n} x^{opt}_{i,k,z}c_i)\geq \rho$. We next construct two assortments based on $\mathbf{x}^{opt}$: The first assortment, denoted by $\mathbf{x}^{opt}_{\leq  t_\rho}$, is constructed by removing all products displayed after $t_\rho$ from the optimal solution. The second assortment, denoted by $\mathbf{x}^{opt}_{> t_\rho}$, is constructed by removing all products scheduled earlier than $t_\rho$ from the optimal solution, then ``shifting'' the rest of products $t_\rho$ stages ahead.  It is clear that the reachability of every product in the first assortment is no smaller than $\rho$.

We first prove the following inequality:
\begin{equation}
\label{Eq:0a}
\rho f(\mathbf{x}^{opt}_{>  t_\rho})\geq  f(\mathbf{x}^{opt})-f(\mathbf{x}^{opt}_{\leq t_\rho})
\end{equation}

According to the definition of $\mathbf{x}^{opt}_{\leq  t_\rho}$, we can derive the expected revenue of the optimal solution $\mathbf{x}^{opt}$ as follows:
\begin{eqnarray}
f(\mathbf{x}^{opt}) 
&=& f(\mathbf{x}^{opt}_{\leq t_\rho})\\
&& + \sum\limits_{t=t_\rho+1}^m \frac{F(\sum\limits_{z=1}^{t-1}\sum\limits_{k=1}^w\sum\limits_{i=1}^{n} x^{opt}_{i,k,z}c_i) (\sum\limits_{k=1}^w \sum\limits_{i=1}^{n}r_i\beta_{i,k}x^{opt}_{i,k,t})}{(1+\sum\limits_{z=1}^{t-1}\sum\limits_{k=1}^w\sum\limits_{i=1}^{n}{\beta}_{i,k}x^{opt}_{i,k,z})(1+\sum\limits_{z=1}^{t}\sum\limits_{k=1}^w\sum\limits_{i=1}^{n}{\beta}_{i,k}x^{opt}_{i,k,z})}\label{eq:0c}
\end{eqnarray}

 We next analyze the expected revenue of $\mathbf{x}^{opt}_{>  t_\rho}$. For ease of presentation, let $\sigma_i$ denote the number of exposures of product $i$ in the optimal solution before stage $t_\rho$. For ease of presentation, define $\sum\limits_{z=t_{\rho}+1}^{t_\rho}\sum\limits_{k=1}^w\sum\limits_{i=1}^{n}  x^{opt}_{i,k,z}c_i =0$ and $\sum\limits_{z=t_\rho+1}^{t_\rho}\sum\limits_{k=1}^w\sum\limits_{i=1}^{n} {\beta}_{i,k-\sigma_i}x^{opt}_{i,k,z}=0$.
\begin{eqnarray}
f(\mathbf{x}^{opt}_{>  t_\rho}) &=& \sum\limits_{t=t_\rho+1}^m \frac{F(\sum\limits_{z=t_{\rho}+1}^{t-1}\sum\limits_{k=1}^w\sum\limits_{i=1}^{n}  x^{opt}_{i,k,z}c_i) (\sum\limits_{k=1}^w \sum\limits_{i=1}^{n}r_i\beta_{i,k-\sigma_i}x^{opt}_{i,k,t})}{(1+\sum\limits_{z=t_\rho+1}^{t-1}\sum\limits_{k=1}^w\sum\limits_{i=1}^{n}{\beta}_{i,k-\sigma_i}x^{opt}_{i,k,z})(1+\sum\limits_{z=t_\rho+1}^{t}\sum\limits_{k=1}^w\sum\limits_{i=1}^{n}{\beta}_{i,k-\sigma_i}x^{opt}_{i,k,z})}~\nonumber\\
&=& \sum\limits_{t=t_\rho+1}^m \frac{F(\sum\limits_{z=t_{\rho}+1}^{t-1} \sum\limits_{k=1}^w \sum\limits_{i=1}^{n} x^{opt}_{i,k,z}c_i) (\sum\limits_{k=1}^w \sum\limits_{i=1}^{n}r_i\beta_{i,k-\sigma_i}x^{opt}_{i,k,t})}{(1+\sum\limits_{z=t_\rho+1}^{t-1}\sum\limits_{k=1}^w \sum\limits_{i=1}^{n}{\beta}_{i,k-\sigma_i}x^{opt}_{i,k,z})(1+\sum\limits_{z=t_\rho+1}^{t}\sum\limits_{k=1}^w \sum\limits_{i=1}^{n}{\beta}_{i,k-\sigma_i}x^{opt}_{i,k,z})} \label{eq:0b}
\end{eqnarray}

To prove inequality (\ref{Eq:0a}), it suffice to prove that the value of (\ref{eq:0c}) is upper bounded by $\rho$ times the value of (\ref{eq:0b}). We next prove a stronger result, that is,  for every $t\in[t_\rho+1, m]$:
\begin{equation}
\begin{split}
 &\frac{F(\sum\limits_{z=1}^{t-1}\sum\limits_{k=1}^w\sum\limits_{i=1}^{n} x^{opt}_{i,k,z}c_i)( \sum\limits_{k=1}^w \sum\limits_{i=1}^{n}r_i\beta_{i,k}x^{opt}_{i,k,t})}{(1+\sum\limits_{z=1}^{t-1}\sum\limits_{k=1}^w\sum\limits_{i=1}^{n}{\beta}_{i,k}x^{opt}_{i,k,z})(1+\sum\limits_{z=1}^{t}\sum\limits_{k=1}^w\sum\limits_{i=1}^{n}{\beta}_{i,k}x^{opt}_{i,k,z})} \\
 &\leq \frac{\rho F(\sum\limits_{z=t_{\rho}+1}^{t-1} \sum\limits_{k=1}^w \sum\limits_{i=1}^{n} x^{opt}_{i,k,z}c_i) (\sum\limits_{k=1}^w \sum\limits_{i=1}^{n}r_i\beta_{i,k-\sigma_i}x^{opt}_{i,k,t})}{(1+\sum\limits_{z=t_\rho+1}^{t-1}\sum\limits_{k=1}^w \sum\limits_{i=1}^{n}{\beta}_{i,k-\sigma_i}x^{opt}_{i,k,z})(1+\sum\limits_{z=t_\rho+1}^{t}\sum\limits_{k=1}^w \sum\limits_{i=1}^{n}{\beta}_{i,k-\sigma_i}x^{opt}_{i,k,z})}
 \end{split}
 \label{eq:0d}
\end{equation}

We first prove that the denominator of LHS of (\ref{eq:0d}) is no smaller than the denominator of RHS of (\ref{eq:0d}). This is true because for all $i\in[n]$, we have $\sum\limits_{z=1}^{t-1}\sum\limits_{k=1}^w{\beta}_{i,k}x^{opt}_{i,k,z}\geq \sum\limits_{z=t_\rho+1}^{t-1}\sum\limits_{k=1}^w {\beta}_{i,k-\sigma_i}x^{opt}_{i,k,z}$ and $\sum\limits_{z=1}^{t}\sum\limits_{k=1}^w {\beta}_{i,k}x^{opt}_{i,k,z}\geq \sum\limits_{z=t_\rho+1}^{t}\sum\limits_{k=1}^w {\beta}_{i,k-\sigma_i}x^{opt}_{i,k,z}$.

We next focus on proving that
\begin{equation*}
F(\sum\limits_{z=1}^{t-1} \sum\limits_{k=1}^w \sum\limits_{i=1}^{n}x^{opt}_{i,k,z}c_i) (\sum\limits_{k=1}^w \sum\limits_{i=1}^{n}r_i\beta_{i,k}x^{opt}_{i,k,t} )
\leq \rho F(\sum\limits_{z=t_{\rho}+1}^{t-1}\sum\limits_{k=1}^w \sum\limits_{i=1}^{n} x^{opt}_{i,k,z}c_i) (\sum\limits_{k=1}^w \sum\limits_{i=1}^{n}r_i\beta_{i,k-\sigma_i}x^{opt}_{i,k,t})
\end{equation*}

Due to Assumption \ref{as:!!}, we have $\beta_{i,k}\leq \beta_{i,k-\sigma_i}$ for every $i\in [n]$ and $k>\sigma_i$, it follows that for all $t\in[m]$, we have $\sum\limits_{k=1}^w\sum\limits_{i=1}^{n}r_i\beta_{i,k}x^{opt}_{i,k,t} \leq \sum\limits_{k=1}^w\sum\limits_{i=1}^{n}r_i\beta_{i,k-\sigma_i}x^{opt}_{i,k,t}$. Moreover, due to Assumption \ref{as:!}, the following inequality holds for every $t\in[t_\rho+1, m]$:
\begin{eqnarray}
\label{eq:999}
F(\sum\limits_{z=t_{\rho}+1}^{t-1} \sum\limits_{k=1}^w\sum\limits_{i=1}^{n} x^{opt}_{i,k,z}c_i) \geq F(\sum\limits_{z=1}^{t-1}\sum\limits_{k=1}^w\sum\limits_{i=1}^{n}x^{opt}_{i,k,z}c_i) / F(\sum\limits_{z=1}^{t_\rho} \sum\limits_{k=1}^w\sum\limits_{i=1}^{n} x^{opt}_{i,k,z}c_i)\end{eqnarray}

Inequality (\ref{eq:999}) together with the assumption that  $F(\sum\limits_{z=1}^{t_\rho} \sum\limits_{k=1}^w\sum\limits_{i=1}^{n} x^{opt}_{i,k,z}c_i)\leq \rho$ implies the following inequality:
\begin{eqnarray*}
F(\sum\limits_{z=t_{\rho}+1}^{t-1} \sum\limits_{k=1}^w\sum\limits_{i=1}^{n} x^{opt}_{i,k,z}c_i)\geq F(\sum\limits_{z=1}^{t-1}\sum\limits_{k=1}^w\sum\limits_{i=1}^{n}x^{opt}_{i,k,z}c_i) / F(\sum\limits_{z=1}^{t_\rho} \sum\limits_{k=1}^w\sum\limits_{i=1}^{n} x^{opt}_{i,k,z}c_i) \geq  F(\sum\limits_{z=1}^{t-1}\sum\limits_{k=1}^w\sum\limits_{i=1}^{n}x^{opt}_{i,k,z}c_i)/\rho
\end{eqnarray*}


 This finishes the proof of (\ref{eq:0d}), which implies (\ref{Eq:0a}), that is, $\rho f(\mathbf{x}^{opt}_{>  t_\rho})\geq  f(\mathbf{x}^{opt})-f(\mathbf{x}^{opt}_{\leq t_\rho})$. Now we are ready to put it all together. Because $\mathbf{x}^{opt}$ is the optimal solution, we have
 $ f(\mathbf{x}^{opt}_{>  t_\rho})\leq f(\mathbf{x}^{opt})$. Together with (\ref{Eq:0a}), we have $f(\mathbf{x}^{opt}_{\leq t_\rho})\geq (1-\rho)f(\mathbf{x}^{opt})$. According to the definition of $\mathbf{x}^{opt}_{\leq t_\rho}$, the reachability of all products in $\mathbf{x}^{opt}_{\leq t_\rho}$ is at least $\rho$. Thus, $\mathbf{x}^{opt}_{\leq t_\rho}$ is such a solution as specified in Lemma \ref{lem:1}. $\Box$

\section{Approximate Solution}

In this section, we develop an approximate solution to our problem. For ease of presentation, given any $\mathbf{x}\in \mathcal{X}$, define
   \[g(\mathbf{x})=\sum\limits_{t=1}^m \frac{\sum\limits_{k=1}^w \sum\limits_{i=1}^{n}r_i\beta_{i,k}x_{i,k,t}}{(1+\sum\limits_{z=1}^{t-1}\sum\limits_{k=1}^w \sum\limits_{i=1}^{n}{\beta}_{i,k}x_{i,k,z})(1+\sum\limits_{z=1}^{t}\sum\limits_{k=1}^w \sum\limits_{i=1}^{n}{\beta}_{i,k}x_{i,k,z})}\]

Note that $g(\mathbf{x})$ is the expected revenue of $\mathbf{x}$ when the reachability of all stages are 1, e.g., this happens when the patience budget of the consumer is always infinity. Before presenting our algorithm, we first introduce a new problem $\textbf{P.1}$ whose solution is a key ingredient of algorithm.
   \begin{center}
   \framebox[0.45\textwidth][c]{
\enspace
\begin{minipage}[t]{0.45\textwidth}
\small
$\textbf{P.1}$
\emph{Maximize$_{\mathbf{x} \in \mathcal{X} }$ $g(\mathbf{x})$}\\
\textbf{subject to:} $
F(\sum\limits_{z=1}^{m} \sum\limits_{k=1}^w \sum\limits_{i=1}^{n} x^{opt}_{i,k,z}c_i)\geq \rho \quad \mbox{(C1)}$
\end{minipage}
}
\end{center}
\vspace{0.1in}

The objective of $\textbf{P.1}$ is to identify the best feasible assortment $\mathbf{x}$ that maximizes $g(\mathbf{x})$ subject to (C1). The condition (C1) ensures that the reachability of all non-empty stages must be no smaller than $\rho$. As compared with the original problem $\textbf{P.0}$, we move the variables of patience cost from the objective function to the constraint (C1) in $\textbf{P.1}$, making it approachable.

Given the formulation of  $\textbf{P.1}$, we are now ready to present our algorithm, called \emph{Assortment Optimization under Cascade Multinomial Logit model} (ACME),  for finding an approximate solution to  \textbf{P.0}.

\textbf{Description of ACME.}
\begin{enumerate}
\item Solve $\textbf{P.1}$ approximately and get a solution $\mathbf{x}'$.
\item Solve \textbf{P.0} with $m=1$ optimally and get a solution $\mathbf{x}{''}$.
\item Return the better solution between $\mathbf{x}{'}$ and  $\mathbf{x}{''}$ as the final solution.
\end{enumerate}

We first discuss the second step of ACME. It was worth noting that when there is only one stage, e.g, $m=1$, \textbf{P.0} is reduced to the classic assortment optimization problem subject to a cardinality constraint. We can solve it optimally  based on \citep{rusmevichientong2010dynamic} and obtain $\mathbf{x}{''}$. We next present the main theorem of this paper. It says that if we can find an approximation algorithm for  $\textbf{P.1}$, then we can solve \textbf{P.0} approximately.

\begin{theorem}
\label{thm:1}
If there exists an $\kappa$-approximate solution to $\textbf{P.1}$, then for any $\rho\in[0,1]$, ACME achieves  $\frac{\kappa\rho(1-\rho)}{2}$ approximation ratio  to $\textbf{P.0}$.
\end{theorem}
\emph{Proof:} Recall that $t_\rho$ is the last stage in $\mathbf{x}^{opt}$ whose reachability is no smaller than $\rho$, e.g.,  $t_\rho =\arg\max_{t\in[m]} F(\sum\limits_{z=1}^{t-1} \sum\limits_{k=1}^w \sum\limits_{i=1}^{n} x^{opt}_{i,k,z}c_i)\geq \rho$.  We first prove that $f(\mathbf{x}^{opt}_{\leq t_\rho})\leq g(\mathbf{x}^*) + f(\mathbf{x}'')$, where  $\mathbf{x}^*$ denotes the optimal solution to $\textbf{P.1}$. Let $\mathbf{x}^{opt}_{<  t_\rho}$ denote a ``sub'' schedule of $\mathbf{x}^{opt}$, removing all products scheduled after $t_\rho-1$ from $\mathbf{x}^{opt}$.
\begin{eqnarray}
f(\mathbf{x}^{opt}_{\leq t_\rho})&=& f(\mathbf{x}^{opt}_{< t_\rho})+ \frac{F(\sum\limits_{z=1}^{t_\rho-1} \sum\limits_{k=1}^w \sum\limits_{i=1}^{n} x^{opt}_{i,k,z}c_i) (\sum\limits_{i=1}^{n}\sum\limits_{k=1}^w r_i\beta_{i,k}x^{opt}_{i,k,t_\rho})}{(1+\sum\limits_{z=1}^{t_\rho-1}\sum\limits_{i=1}^{n}{\beta}_{i,k}x^{opt}_{i,k,z})(1+\sum\limits_{z=1}^{t_\rho}\sum\limits_{k=1}^w \sum\limits_{i=1}^{n}{\beta}_{i,k}x^{opt}_{i,k,z})}\\
&\leq& g(\mathbf{x}^{opt}_{< t_\rho})+\frac{\sum\limits_{i=1}^{n}\sum\limits_{k=1}^w r_i\beta_{i,k}x^{opt}_{i,k,t_\rho}}{(1+\sum\limits_{z=1}^{t_\rho-1}\sum\limits_{k=1}^w \sum\limits_{i=1}^{n}{\beta}_{i,k}x^{opt}_{i,k,z})(1+\sum\limits_{z=1}^{t_\rho}\sum\limits_{k=1}^w \sum\limits_{i=1}^{n}{\beta}_{i,k}x^{opt}_{i,k,z})}\label{eq:3}\\
&\leq& g(\mathbf{x}^*)+\frac{\sum\limits_{i=1}^{n}\sum\limits_{k=1}^w r_i\beta_{i,1}x^{opt}_{i,k,t_\rho}}{1+\sum\limits_{z=1}^{t_\rho}\sum\limits_{k=1}^w \sum\limits_{i=1}^{n}{\beta}_{i,k}x^{opt}_{i,k,z}}\label{eq:4}\\
&\leq& g(\mathbf{x}^*)+\frac{\sum\limits_{i=1}^{n}\sum\limits_{k=1}^w r_i\beta_{i,1}x^{opt}_{i,k,t_\rho}}{1+\sum\limits_{i=1}^{n}\sum\limits_{k=1}^w {\beta}_{i,1}x^{opt}_{i,k,t_\rho}}\label{eq:1}\\
&\leq& g(\mathbf{x}^*) + f(\mathbf{x}'')\label{eq:2}
\end{eqnarray}
Inequality (\ref{eq:3}) is due to $f(\mathbf{x}^{opt}_{< t_\rho})\leq g(\mathbf{x}^{opt}_{< t_\rho})$, and $F(\sum\limits_{z=1}^{t_\rho-1} \sum\limits_{k=1}^w \sum\limits_{i=1}^{n} x^{opt}_{i,k,z}c_i) \leq 1$. Inequality (\ref{eq:4}) is due to $\mathbf{x}^{opt}_{\leq t_\rho}$ is a feasible solution to $\textbf{P.1}$ and $\mathbf{x}^*$ is the optimal solution to $\textbf{P.1}$,  $\beta_{i,1}\geq \beta_{i,k}$ for all $i\in [n]$, and $(1+\sum\limits_{z=1}^{t_\rho-1}\sum\limits_{k=1}^w \sum\limits_{i=1}^{n}{\beta}_{i,k}x^{opt}_{i,k,z})>1$. Inequality (\ref{eq:1}) is due to $\sum\limits_{z=1}^{t_\rho}\sum\limits_{k=1}^w \sum\limits_{i=1}^{n}{\beta}_{i,k}x^{opt}_{i,k,z} \geq \sum\limits_{i=1}^{n}\sum\limits_{k=1}^w {\beta}_{i,1}x^{opt}_{i,k,t_\rho}$. Note that the second term of (\ref{eq:1}) can be viewed as the expected revenue of the following single stage assortment: for each product $i$, selecting $i$ if $x^{opt}_{i,k,t_\rho}=1$. Recall that $\mathbf{x}''$ is the optimal solution to the single stage assortment optimization problem, thus inequality (\ref{eq:2}) holds.

Assume $g(\mathbf{x}')\geq \kappa g(\mathbf{x}^*)$, based on inequality (\ref{eq:2}), we have $f(\mathbf{x}^{opt}_{\leq t_\rho})\leq g(\mathbf{x}')/\kappa+f(\mathbf{x}'')$. Because the reachability of all stages under $\mathbf{x}'$ is lower bounded by $\rho$, we have $f(\mathbf{x}')\geq \rho g(\mathbf{x}')$. It follows that $f(\mathbf{x}^{opt}_{\leq t_\rho})\leq f(\mathbf{x}')/\kappa\rho+f(\mathbf{x}'')$. Together with (\ref{Eq:0a}), we have $f(\mathbf{x}')/\kappa\rho+f(\mathbf{x}'')\geq (1-\rho)f(\mathbf{x}^{opt})$.  It follows that $\max\{f(\mathbf{x}'), f(\mathbf{x}''\}\geq \frac{\kappa\rho(1-\rho)}{2}f(\mathbf{x}^{opt})$. Because ACME picks the better one between $\mathbf{x}'$ and $\mathbf{x}''$ as the final solution, this theorem holds. $\Box$

In the next subsection, we propose a solution to $\textbf{P.1}$ based on  dynamic programming, and we prove in Lemma \ref{lem:14} that it achieves a $ \frac{1-\epsilon(1+\epsilon)}{(1+\epsilon(1+\epsilon))^2} $ approximation ratio for any $\epsilon>0$. By setting $\kappa=\frac{1-\epsilon(1+\epsilon)}{(1+\epsilon(1+\epsilon))^2} $ in Theorem \ref{thm:1}, we have the following performance bound for ACME.

\begin{corollary}
Given that we develop a $ \frac{1-\epsilon(1+\epsilon)}{(1+\epsilon(1+\epsilon))^2} $-approximate solution to $\textbf{P.1}$ for any $\epsilon>0$, ACME achieves a $\frac{ \rho(1-\rho)(1-\epsilon(1+\epsilon))}{2(1+\epsilon(1+\epsilon))^2}$ approximation ratio to \textbf{P.0}  for any $\rho\in[0,1]$.
\end{corollary}

The rest of this paper is devoted to developing a $ \frac{1-\epsilon(1+\epsilon)}{(1+\epsilon(1+\epsilon))^2} $-approximate solution to $\textbf{P.1}$ based on dynamic programming.  We build our solution on the recent advances in the assortment optimization problem subject to one capacity constraint \citep{desir2014near}, we generalize their idea and provide an approximate algorithm for the assortment optimization problem subject to a capacity constraint, a cardinality constraint, and a partition matroid-type feasibility  constraint (the same product can be displayed at
most once in each stage).

\subsection{A Dynamic Programming based Solution to $\textbf{P.1}$}
\label{sec:sub}
Before presenting our solution, we first introduce some notations.
Define $\beta_{\min}=\min_{i\in [n], k\in[w]} \beta_{i,k}$ and $\beta_{\max}=\max_{i\in [n], k\in[w]} \beta_{i,k}$. Let $\gamma_{i,k}=r_i\beta_{i,k}$, define $\gamma_{\min}=\min_{i\in [n], k\in[w]} \gamma_{i,k}$, and $\gamma_{\max}=\max_{i\in [n], k\in[w]} \gamma_{i,k}$.

For a given $\epsilon>0$, we first construct a geometric grid $I\times J$ where $I$ and $J$ are defined as follows.
\[I=\{\gamma_{\min}(1+\epsilon)^a\mid a\in [ \lceil \ln\frac{d\gamma_{\max}}{\epsilon\gamma_{\min}}\rceil]\}, J=\{\beta_{\min}(1+\epsilon)^b\mid b\in [\lceil\ln\frac{d\beta_{\max}}{\epsilon\beta_{\min}}\rceil]\}\]
Then we build a group of guesses $\mu=\{\mu_1, \mu_2, \ldots, \mu_m\} \in I^m$ and  $\nu=\{\nu_1,\nu_2, \ldots, \nu_m\}\in J^m$.
We go through all guesses $(\mu, \nu)\in I^m \times J^m$ and check whether or not there exists a solution $\mathbf{x}\in \mathcal{X}$ such that $\sum\limits_{k=1}^w \sum\limits_{i=1}^{n}\gamma_{i,k} x_{i,k,z}$ is approximately equal to $\mu_z$ and $\sum\limits_{k=1}^w \sum\limits_{i=1}^{n} \beta_i x_{i,k,z}$ is approximately equal to $\nu_z$ for all $z\in[m]$.

For a given guess $(\mu, \nu)\in I^m \times J^m$, we discretize the values of $\gamma_{i,k}$ and $\beta_{i,k}$, and define $\tilde{\gamma}_{i,k,z}$ and $\tilde{\beta}_{i,k,z}$ for all $i\in [n]$, $k\in[w]$, $z\in[m]$ as follows:
\[\tilde{\gamma}_{i,k,z}=\lceil\frac{\gamma_{i,k}}{\mu_z\epsilon/d}\rceil, \tilde{\beta}_{i,k,z}=\lfloor\frac{\beta_{i,k}}{\nu_z\epsilon/d}\rfloor\]
Note that when $\mu_z\geq \gamma_{i,k}$ and $\nu_z\geq \beta_{i,k}$ for all $z\in[m], i\in[n], k\in[w]$, we have $\sum_{i\in[n], k\in[w]}\tilde{\gamma}_{i,k,z} \leq d\times \lceil d/\epsilon\rceil\leq d(d/\epsilon +1)$ and $\sum_{i\in[n], k\in[w]}\tilde{\beta}_{i,k,z} \leq d\times \lfloor d/\epsilon\rfloor\leq d(d/\epsilon +1)$ for all $z\in [m]$.

 Denote by function $h(j, \mathbf{u}, \mathbf{v}, \mathbf{l})$ for $(j, \mathbf{u}, \mathbf{v}, \mathbf{l}) \in [n]\times [d(d/\epsilon+1)]^m \times [d(d/\epsilon+1)]^m \times [d]^m$ the optimal solution value of the following problem:
\begin{equation*}
\begin{split}
&h(j, \mathbf{u}, \mathbf{v}, \mathbf{l}):= \min_{\mathbf{x}\in \mathcal{X}}\{\sum\limits_{z=1}^{m}\sum\limits_{k=1}^{w}\sum\limits_{i=1}^{j}c_ix_{i,k,z}: \\
&\sum\limits_{z=1}^{m}\sum\limits_{k=1}^{w}\sum\limits_{i=1}^{j}\tilde{\gamma}_{i,k,z}x_{i,k,z}=u_z, \sum\limits_{z=1}^{m}\sum\limits_{k=1}^{w}\sum\limits_{i=1}^{j}\tilde{\beta}_{i,k,z}x_{i,k,z}=v_z, \sum\limits_{z=1}^{m}\sum\limits_{k=1}^{w}\sum\limits_{i=1}^{j}x_{i,k,z}=l_z\}
\end{split}
\end{equation*} where $u_z$ is the $z$-th component of $\mathbf{u}$, $v_z$ is the $z$-th component of $\mathbf{v}$, and $l_z$ the $z$-the component of $\mathbf{l}$. Intuitively,  $h(j, \mathbf{u}, \mathbf{v}, \mathbf{l})$ represents the minimum total patience cost of any assortment of products $\{1, 2, \ldots, j\}$ such that $\sum\limits_{z=1}^{m}\sum\limits_{k=1}^{w}\sum\limits_{i=1}^{j}\tilde{\gamma}_{i,k,z}x_{i,k,z}=u_z, \sum\limits_{z=1}^{m}\sum\limits_{k=1}^{w}\sum\limits_{i=1}^{j}\tilde{\beta}_{i,k,z}x_{i,k,z}=v_z, \sum\limits_{z=1}^{m}\sum\limits_{k=1}^{w}\sum\limits_{i=1}^{j}x_{i,k,z}=l_z\}$.  We set the initial values as follows: we first set $h(j, \mathbf{u}, \mathbf{v}, \mathbf{l})=+\infty$ when $j=0$  or there exists some $z\in[m]$ such that $u_z<0$ or  $v_z<0$ or $l_z<0$, and then set $h(j, \mathbf{u}, \mathbf{v}, \mathbf{l})=0$ when the following conditions are satisfied: $j=0$, and for all $z\in[m]$, $u_z=0$, $v_z=0$, and $l_z=0$.


For ease of presentation, we next introduce an alternative way to represent the schedule of a product: For every product $i\in [n]$, we use a binary vector $\mathbf{y}_i=\{y_{i1}, \cdots, y_{im}\}\in\{0,1\}^m$ to represent the schedule of $i\in[n]$  such that $y_{iz}=1$ if $i$ is displayed in stage $z$, and $y_{iz}=0$ otherwise.
Given a schedule $\mathbf{y}_j$ of $j$, assume the index of the $k$-th non-zero element is $z_k$. Define $\mathbf{y}_j\tilde{\gamma}_j$   as a vector that replaces the $k$-th non-zero element of $\mathbf{y}_j$ with $\tilde{\gamma}_{j,k,z_k}$  for all $k\in [w]$, and define $\mathbf{y}_j\tilde{\beta}_j$ as a vector that replaces the $k$-th non-zero element of $\mathbf{y}_j$ with  $\tilde{\beta}_{j,k,z_k}$ for all $k\in [w]$. Let $|\mathbf{y}_j|_1$ denote the $L_1$ norm of $\mathbf{y}_j$, then we fill up the dynamic program table using the following recurrence function:
\begin{equation*}
h(j, \mathbf{u}, \mathbf{v}, \mathbf{l})=
\min_{\mathbf{y}_j}
h(j-1, \mathbf{u}-\mathbf{y}_j\tilde{\gamma}_j, \mathbf{v}-\mathbf{y}_j\tilde{\beta}_j, \mathbf{l}-\mathbf{y}_j)+|\mathbf{y}_j|_1c_j
\end{equation*}

 One way to compute $h(j, \mathbf{u}, \mathbf{v}, \mathbf{l})$ is to enumerate all possible schedules $\mathbf{y}_j$ of $j$ and find the one that minimizes $h(j-1, \mathbf{u}-\mathbf{y}_j\tilde{\gamma}_j, \mathbf{v}-\mathbf{y}_j\tilde{\beta}_j, \mathbf{l}-\mathbf{y}_j)+|\mathbf{y}_j|_1c_j$. Because each product can only be displayed at most $w$ times and there are $m$ stages, the time complexity of enumerating all $\mathbf{y}_j$ is $O(m^w)$.
\begin{lemma}
The time complexity of the dynamic program is $O(m^w (d(d/\epsilon +1))^{2m} \ln\frac{d\gamma_{\max}}{\epsilon\gamma_{\min}}\ln\frac{d\beta_{\max}}{\epsilon\beta_{\min}})$.
\end{lemma}
\emph{Proof:}
Our proof is based on the following three observations. First, the total number $|I|\cdot|J|$ of guesses  is bounded by $=O(\ln\frac{d\gamma_{\max}}{\epsilon\gamma_{\min}}\ln\frac{d\beta_{\max}}{\epsilon\beta_{\min}})$. Second, enumerating all $(\mathbf{u},  \mathbf{v})\in  [d(d/\epsilon+1)]^m \times [d(d/\epsilon+1)]^m $ requires time complexity of $(d(d/\epsilon +1))^{2m}$.  Third, the time complexity of computing $h(j, \mathbf{u}, \mathbf{v}, \mathbf{l})$ is $O(m^w)$, e.g., this is done by enumerating all possible $\mathbf{y}_j$. It follows that the total time complexity of the dynamic program is $O(m^w (d(d/\epsilon +1))^{2m} \ln\frac{d\gamma_{\max}}{\epsilon\gamma_{\min}}\ln\frac{d\beta_{\max}}{\epsilon\beta_{\min}})$. $\Box$

Note that the running time of the dynamic program  increases exponential with the number of stages $m$. It would be important to develop effective algorithms for large $m$, which we leave for future work.

We next prove that the dynamic program is a $\frac{1-\epsilon(1+\epsilon)}{(1+\epsilon(1+\epsilon))^2}$ approximate solution to $\textbf{P.1}$.
\begin{lemma}
\label{lem:14}
Let $\mathbf{x}^*$ denote the optimal solution to $\textbf{P.1}$. Recall that we use $\mathbf{x}'$ to denote the solution returned from the dynamic program. For any $\epsilon>0$, we have $g(\mathbf{x}')\geq \frac{1-\epsilon(1+\epsilon)}{(1+\epsilon(1+\epsilon))^2} g(\mathbf{x}^*)$.
\end{lemma}
\emph{Proof:} Let $m^*=\arg\max_z (\sum\limits_{k=1}^{w}\sum\limits_{i=1}^{n}{\gamma}_{i,k,z}x^*_{i,k,z}\neq0)$, e.g., the optimal solution $\mathbf{x}^*$ only utilizes the first $m^*$ stages\footnote{It is easy to show that there is an optimal assortment that does not contain ``gaps'' between
stages. Otherwise, we can remove those gaps by shifting all products ahead such that the expected revenue does not decrease.}. Assume  for all $z\in [m^*]$, $\gamma_{\min}(1+\epsilon)^{a_z} \leq\sum\limits_{k=1}^{w} \sum\limits_{i=1}^{n}{\gamma}_{i,k,z}x^*_{i,k,z} \leq \gamma_{\min}(1+\epsilon)^{a_z+1}$ and $\beta_{\min}(1+\epsilon)^{b_z} \leq \sum\limits_{k=1}^{w} \sum\limits_{i=1}^{n}{\beta}_{i,k}x^*_{i,k,z} \leq \beta_{\min}(1+\epsilon)^{b_z+1}$. Recall that the dynamic program enumerates all guesses in $I^m\times J^m$. Consider the case when $\{(\gamma_{\min}(1+\epsilon)^{a_z+1},\beta_{\min}(1+\epsilon)^{b_z+1})\mid z\in[m^*]\}$ is enumerated, let $u_z^*=\sum\limits_{k=1}^{w}\sum\limits_{i=1}^{n}\tilde{\gamma}_{i,k,z}x^*_{i,k,z}$, $v_z^*=\sum\limits_{k=1}^{w}\sum\limits_{i=1}^{n}\tilde{\beta}_{i,k,z}x^*_{i,k,z}$, and $l_z^*=\sum\limits_{k=1}^{w}\sum\limits_{i=1}^{n}x^*_{i,k,z}$ denote the summation of the scaled values of the optimal solution for all $z\in[m^*]$. It is clear that $h(n, \mathbf{u}^*, \mathbf{v}^*, \mathbf{l}^*)\leq \rho$ where the $z$-th component of $\mathbf{u}^*$ is $u^*_z$, the $z$-th component of $\mathbf{v}^*$ is $v^*_z$, and the $z$-th component of $\mathbf{l}^*$ is $l^*_z$.

We first give a lower bound on  $\sum\limits_{k=1}^{w}\sum\limits_{i=1}^{n}\gamma_{i,k}x'_{i,k,z}$ for all $z\in[m^*]$,
\begin{eqnarray}
&&\sum\limits_{k=1}^{w}\sum\limits_{i=1}^{n}\gamma_{i,k}x'_{i,k,z}\geq  \sum\limits_{k=1}^{w}\sum\limits_{i=1}^{n}\tilde{\gamma}_{i,k,z}x^*_{i,k,z} \epsilon\gamma_{\min}(1+\epsilon)^{a_z+1}/d-\epsilon\gamma_{\min}(1+\epsilon)^{a_z+1}\\
&=& u_z^*\epsilon\gamma_{\min}(1+\epsilon)^{a_z+1}/d -\epsilon\gamma_{\min}(1+\epsilon)^{a_z+1}\\
&\geq& u_z^*\epsilon\gamma_{\min}(1+\epsilon)^{a_z+1}/d-\epsilon(1+\epsilon)\sum\limits_{k=1}^{w} \sum\limits_{i=1}^{n}\gamma_{i,k}x^*_{i,k,z}\\
&\geq& \sum\limits_{k=1}^{w}\sum\limits_{i=1}^{n}\gamma_{i,k}x^*_{i,k,z} -\epsilon(1+\epsilon) \sum\limits_{k=1}^{w}\sum\limits_{i=1}^{n}\gamma_{i,k}x^*_{i,k,z}=(1-\epsilon(1+\epsilon))\sum\limits_{k=1}^{w} \sum\limits_{i=1}^{n}\gamma_{i,k}x^*_{i,k,z} \label{eq:lastlast}
 \end{eqnarray} where the second inequality is due to the assumption that $\gamma_{\min}(1+\epsilon)^{a_z} \leq \sum\limits_{i=1}^{n}{\gamma}_{i,k,z}x^*_{i,k,z}$ and the last inequality is due to $\tilde{\gamma}_{i,k,z}\geq \frac{\gamma_{i,k}}{\gamma_{\min}(1+\epsilon)^{a_z+1}\epsilon/d}$ for all $i\in [n], k\in[w], z\in [m^*]$.

Then we give an upper bound on  $\sum\limits_{k=1}^{w}\sum\limits_{i=1}^{n}\beta_{i,k}x'_{i,k,z}$ for all $z\in[m^*]$,
\begin{eqnarray}&&\sum\limits_{k=1}^{w} \sum\limits_{i=1}^{n}\beta_{i,k}x'_{i,k,z}
\leq \sum\limits_{k=1}^{w}\sum\limits_{i=1}^n \tilde{\beta}_{i,k,z}x^*_{i,k,z} \epsilon\beta_{\min}(1+\epsilon)^{b_z+1}/d+\epsilon\beta_{\min}(1+\epsilon)^{b_z+1} \\
&=& v_z^*\epsilon\beta_{\min}(1+\epsilon)^{b_z+1}/d+\epsilon\beta_{\min}(1+\epsilon)^{b_z+1}\\
&\leq& v_z^*\epsilon\beta_{\min}(1+\epsilon)^{b_z+1}/d+\epsilon(1+\epsilon)\sum\limits_{k=1}^{w}\sum\limits_{i=1}^{n}{\beta}_{i,k}x^*_{i,k,z} \\
&\leq& \sum\limits_{k=1}^{w}\sum\limits_{i=1}^{n}\beta_{i,k}x^*_{i,k,z}+\epsilon(1+\epsilon)\sum\limits_{k=1}^{w}\sum\limits_{i=1}^{n}{\beta}_{i,k}x^*_{i,k,z}=(1+\epsilon(1+\epsilon))\sum\limits_{k=1}^{w}\sum\limits_{i=1}^{n}{\beta}_{i,k}x^*_{i,k,z} \label{eq:last}
 \end{eqnarray} where the second inequality is due to $\beta_{\min}(1+\epsilon)^{b_z} \leq \sum\limits_{i=1}^{n}{\beta}_{i,k}x^*_{i,k,z}$ and the last inequality is due to $\tilde{\beta}_{i,k,z}\leq  \frac{{\beta}_{i,k}}{\beta_{\min}(1+\epsilon)^{a_z+1}\epsilon/d}$ for all $i\in [n], k\in[w], z\in [m^*]$.

Define $x^*_{i,k,0}=0$ and $x'_{i,k,0}=0$ for all $i\in[n]$ and $k\in[w]$. Based on (\ref{eq:lastlast}) and (\ref{eq:last}), we have
\begin{eqnarray*}
g(\mathbf{x}')&=&\sum\limits_{t=1}^{m^*} \frac{\sum\limits_{k=1}^{w}\sum\limits_{i=1}^{n}\gamma_{i,k}x'_{i,k,t}}{(1+\sum\limits_{z=1}^{t-1}\sum\limits_{k=1}^{w}\sum\limits_{i=1}^{n}{\beta}_{i,k}x'_{i,k,z})(1+\sum\limits_{z=1}^{t}\sum\limits_{k=1}^{w}\sum\limits_{i=1}^{n}{\beta}_{i,k}x'_{i,k,z})}\\
&\geq& \sum\limits_{t=1}^{m^*} \frac{(1-\epsilon(1+\epsilon))\sum\limits_{k=1}^{w}\sum\limits_{i=1}^{n}\gamma_{i,k}x^*_{i,k,t}}{(1+(1+\epsilon(1+\epsilon))\sum\limits_{z=1}^{t-1}\sum\limits_{k=1}^{w}\sum\limits_{i=1}^{n}{\beta}_{i,k}x^*_{i,k,z})(1+(1+\epsilon(1+\epsilon))\sum\limits_{z=1}^{t}\sum\limits_{k=1}^{w}\sum\limits_{i=1}^{n}{\beta}_{i,k}x^*_{i,k,z})}\\
&\geq& \frac{1-\epsilon(1+\epsilon)}{(1+\epsilon(1+\epsilon))^2} \sum\limits_{t=1}^{m^*} \frac{\sum\limits_{k=1}^{w}\sum\limits_{i=1}^{n}\gamma_{i,k}x^*_{i,k,t}}{(1+\sum\limits_{z=1}^{t-1}\sum\limits_{k=1}^{w}\sum\limits_{i=1}^{n}{\beta}_{i,k}x^*_{i,k,z})(1+\sum\limits_{z=1}^{t}\sum\limits_{k=1}^{w}\sum\limits_{i=1}^{n}{\beta}_{i,k}x^*_{i,k,z})}\\
&=& \frac{1-\epsilon(1+\epsilon)}{(1+\epsilon(1+\epsilon))^2} g(\mathbf{x}^*)
\end{eqnarray*}
$\Box$

\section{Conclusion}
In this work, we have considered the assortment optimization problem across multiple stages. Our model allows for both repeated exposures and product-dependent patience cost. We develop an approximation algorithm to this problem whose running time increases exponential with the number of stages. It would be useful to develop effective algorithms when the number of stages is large.
\bibliographystyle{ijocv081}
\bibliography{reference}

\end{document}